\documentclass[10pt,twocolumn,letterpaper]{article}

\usepackage{cvpr}              

\usepackage[dvipsnames]{xcolor}

\usepackage[accsupp]{axessibility}

\definecolor{cvprblue}{rgb}{0.21,0.49,0.74}
\usepackage[pagebackref,breaklinks,colorlinks,citecolor=cvprblue]{hyperref}

\def\paperID{11} 
\def\confName{EarthVision}
\def\confYear{2024}

\title{Deep Generative Data Assimilation in Multimodal Setting}

\author{Yongquan Qu\thanks{Equal contribution; order decided by \texttt{np.random.rand(1)}}\\
Columbia University\\
{\tt\small yq2340@columbia.edu}
\and
Juan Nathaniel\footnotemark[1]\\
Columbia University\\
{\tt\small jn2808@columbia.edu}
\and
Shuolin Li\\
Columbia University\\
{\tt\small sl5487@columbia.edu}
\and
Pierre Gentine\\
Columbia University\\
{\tt\small pg2328@columbia.edu}
}

\begin{document}
\maketitle
\begin{abstract}
Robust integration of physical knowledge and data is key to improve computational simulations, such as Earth system models. Data assimilation is crucial for achieving this goal because it provides a systematic framework to calibrate model outputs with observations, which can include remote sensing imagery and ground station measurements, with uncertainty quantification. Conventional methods, including Kalman filters and variational approaches, inherently rely on simplifying linear and Gaussian assumptions, and can be computationally expensive. Nevertheless, with the rapid adoption of data-driven methods in many areas of computational sciences, we see the potential of emulating traditional data assimilation with deep learning, especially generative models. In particular, the diffusion-based probabilistic framework has large overlaps with data assimilation principles: both allows for conditional generation of samples with a Bayesian inverse framework. These models have shown remarkable success in text-conditioned image generation or image-controlled video synthesis. Likewise, one can frame data assimilation as observation-conditioned state calibration. In this work, we propose \texttt{SLAMS}: Score-based Latent Assimilation in Multimodal Setting. Specifically, we assimilate \emph{in-situ} weather station data and \emph{ex-situ} satellite imagery to calibrate the vertical temperature profiles, globally. Through extensive ablation, we demonstrate that \texttt{SLAMS} is robust even in low-resolution, noisy, and sparse data settings. To our knowledge, our work is the first to apply deep generative framework for multimodal data assimilation using real-world datasets; an important step for building robust computational simulators, including the next-generation Earth system models. Our code is available at: \href{https://github.com/YONGQUAN-QU/SLAMS}{https://github.com/yongquan-qu/SLAMS}.

\end{abstract}    
\section{Introduction}
\label{sec:intro}
\begin{figure}[h!]
    \centering
    \includegraphics[width=0.4\textwidth]{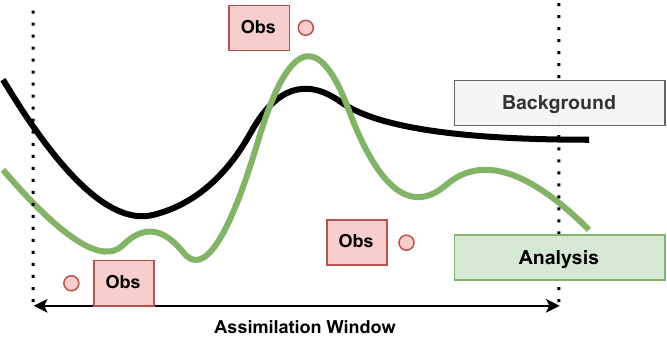}
    \caption{Schematic diagram illustrating the high-level concept of DA. It provides a systematic framework that calibrates background states (model outputs) with multimodal observations (often low-resolution, noisy, and sparse) to produce analysis.}
    \label{fig:slams-da}
\end{figure}

Data assimilation (DA) is crucial for numerous scientific disciplines that require accurate modeling of chaotic systems. It is particularly important across various areas of the geosciences, including fluid dynamics simulations, oceanography, and atmospheric science \cite{carrassi2018data}. At its core, the state estimation problem of DA seeks to calibrate background trajectory of states, \(x^b_{1:T} = (x^b_1, \cdots, x^b_T) \in \mathbb{R}^{T \times N_x}\), given available observations, \(y\), which are often noisy, sparse, and/or multimodal. The calibrated state after assimilation, denoted as \(x^a_{1:T}\), is referred to as the analysis (see Figure \ref{fig:slams-da}). 

\begin{figure*}[h!]
    \centering
    \includegraphics[width=0.7\textwidth]{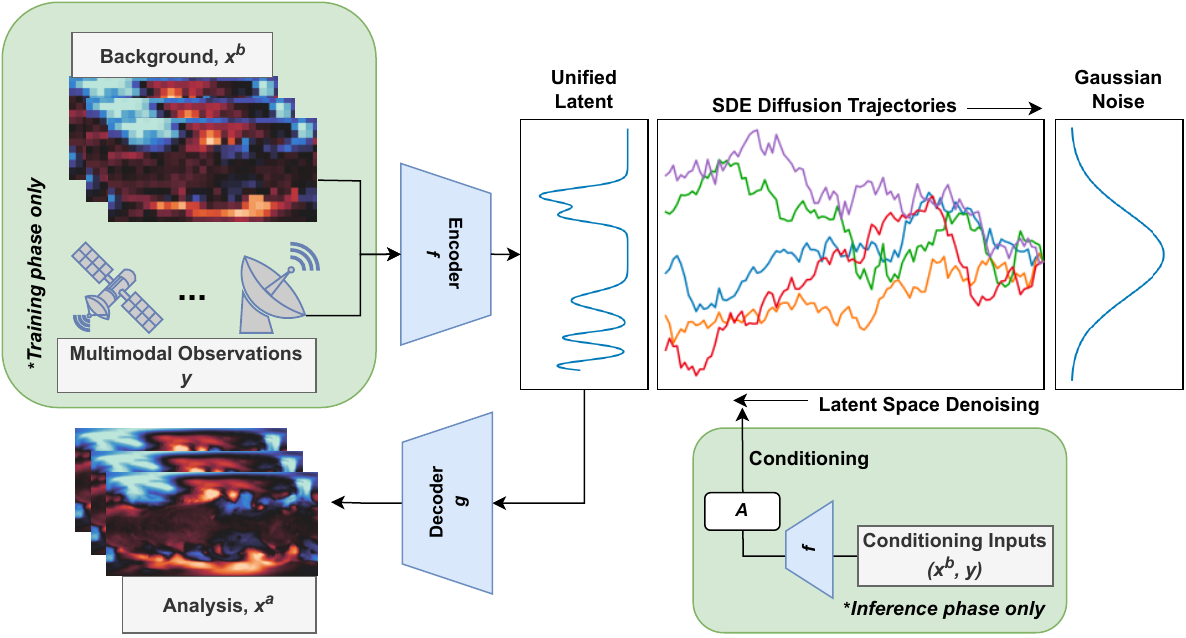}
    \caption{We propose \texttt{SLAMS} - \textbf{S}core-based \textbf{L}atent \textbf{A}ssimilation in \textbf{M}ultimodal \textbf{S}etting. During training, we fit three set of models, including encoder, decoder, and the score network given multimodal data sources (e.g., background states, in-situ sensors, and ex-situ satellite measurements). During inference, we perform latent space denoising through a reverse SDE process by sampling from prior Gaussian distribution and conditioning on encoded background and observations, synthetically coarsified, noisified, and sparsified by a differentiable measurement function $\mathcal{A}$ for ablation purposes.}
    \label{fig:slams-schematic}
\end{figure*}

Conventional DA methods often rely on simplifying assumptions (linearity and Gaussianity) or approximations (tangent linear and adjoint) to formulate closed-form expressions for analysis, such as in Kalman filter-based techniques, or to create a manageable optimization target, as in variational approaches. In many applications, however, the system's underlying dynamics are nonlinear and challenging to encapsulate within a tractable state-space model \cite{revach2022kalmannet}, leading to performance degradation. This necessitates more frequent calibration through assimilation, leading to increased computation cost.

With the recent advances of deep learning for Earth system modeling \cite{reichstein2019deep,kochkov2023neural}, the natural progression is the integration of these data-driven principles into DA for improved state estimation \cite{buizza2022data,farchi2023online}, unresolved scale (parameterization) inference \cite{qu2023can, qu2024joint,brajard2021combining,brajard2021combining, bhouri2022history}, and parameter inference \cite{chen2022autodifferentiable}, either separately or in combination \cite{qu2024joint}. In particular, the rapid adoption of controllable deep generative modeling in text-conditioned image generation \cite{batzolis2021conditional, saharia2022photorealistic} or image-controlled video synthesis \cite{ni2023conditional} could pave the way toward a robust data-driven DA framework. Indeed, the central tenet is similar: we can use observations to condition the generation of analysis. There has been a growing interest to reframe DA through the lens of a deep probabilistic framework \cite{mohd2022deep,bao2022variational,peyron2021latent,cheng2023generalised,arcucci2021deep}, particularly with diffusion as a powerful class of generative model capable of fine-grained controlled generation of high-quality samples \cite{rozet2023score, huang2024diffda}.

However, previous works on data-driven DA used simplified assimilation scenarios: they tend to only assimilate perturbed states as \emph{psuedo-observations} through filtering, noise addition, or sampling \cite{rozet2023score,huang2024diffda}, ignoring actual observations. Observational datasets in real-world are \emph{proxies} without direct link to the target states. For instance, operational forecasting models may assimilate diverse data sources like weather station outputs, satellite imagery, and LiDAR-derived point clouds \cite{lahoz2014data}. Moreover, these approaches tend to rely on computer vision architectures that favor uniform resolution with limited multi-scale processing capabilities, neglecting rich observations with \emph{heterogeneous} spatiotemporal resolutions \cite{xue2014multimodel,mai2022towards, mai2023opportunities}.

To address these challenges, we introduce \texttt{SLAMS} - \textbf{S}core-based \textbf{L}atent \textbf{A}ssimilation in \textbf{M}ultimodal \textbf{S}etting (Figure \ref{fig:slams-schematic}). Fundamentally, \texttt{SLAMS} performs its conditional generative process within a unified latent space. This approach allows for the projection of heterogeneous, multimodal datasets into a common $\mathcal{L}$-dimensional, latent subspace alongside the target states. As a result, it eliminates the cumbersome, assumption-heavy observation operator, $\mathcal{H}: \mathbb{R}^{N_x} \rightarrow \mathbb{R}^{N_y}$, commonly used in traditional DA setup to reconcile heterogeneous spaces \cite{peyron2021latent,cheng2023generalised}. Furthermore, the probabilistic underpinning of \texttt{SLAMS} allows us to generate an ensemble of analysis which naturally facilitates uncertainty quantification. In this study, we adapt score-based latent diffusion model \cite{song2020score,rombach2022high} to calibrate real-world weather states using multimodal observation. Specifically, we assimilate both \emph{in-situ} weather station data and \emph{ex-situ} satellite observations to refine vertical temperature profile in the atmosphere as a proof-of-concept. Our ablation demonstrates that \texttt{SLAMS} produces consistent analysis states, even in low-resolution, noisy, and sparse data settings.
\section{Methodology}
\label{sec:methodology}

Data assimilation at its core attempts to estimate the conditional probability density function of \(p(x_{1:T} \mid y)\)  through a Bayesian inverse formulation as defined in Equation \ref{eq:da},
\begin{equation}
    \label{eq:da}
    p(x_{1:T} \mid y) = \frac{p(y \mid x_{1:T})}{p(y)}\underbrace{p(x_1)\prod_{t=1}^{T-1}p(x_{t+1} \mid x_t)}_{\text{Markovian background prior}}.
\end{equation}

\noindent We elaborate how the right-hand side of Equation \ref{eq:da} could be reconstructed through the scaffolding of score-based diffusion framework. In particular, we discuss our latent version of the model (\texttt{SLAMS}) that enables multimodal assimilation. Finally, we discuss the datasets, experimental setups, and ablation used throughout this work.

\subsection{Multimodality in Unified Latent Space}
Many multimodal observations, $Y^{1:K} = (y^1 \in \mathbb{R}^{T \times M_1}, \cdots, y^K \in \mathbb{R}^{T \times M_K})$, such as cloud cover, vegetation, and radiation tend to be indirectly linked as proxies to the target states, such as vertical temperature. For one, capturing these relationships is non-trivial because $x$ and $Y$ may have different dimensionality and occupy non-overlapping manifolds which render generative models operating directly in the pixel space less effective. Even if we have access to a mapping function, such as the observation operator that resolves the state-observation heterogeneous spaces, $\mathcal{H}:\mathbb{R}^{N_x} \rightarrow \mathbb{R}^{M_1, \cdots, M_K}$, our imperfect knowledge can exponentiate error especially with high-dimensional $K$, and further calibrations to correct biases are required \cite{liang2023machine}. 

Fortunately, with deep learning, we are able to circumvent this problem by learning a sufficiently expressive function that is able to project both the states and observations into a unified, reduced-order $\mathcal{L}$-dimensional latent space. In this work, we define $f_{\theta}: \mathbb{R}^{(N_x + M_1 + \cdots + M_K)} \rightarrow \mathbb{R}^{\mathcal{L}}$ as an encoder that takes as input the concatenation of background state and each observation modality along unique channels for a single timestep. The decoder $g_{\phi}: \mathbb{R}^\mathcal{L} \rightarrow \mathbb{R}^{N_x}$ then maps the latent code back into the pixel-space. 

In the following section, we review previous work on score-based diffusion model and its novel application to DA introduced by \cite{rozet2023score}. In our extension, we define variables $x$ and $y$ to represent the latent representations of the unconditional and conditioning inputs, respectively, including \emph{in-situ} and \emph{ex-situ} observations to align with Equation \ref{eq:da}. 

\subsection{Score-based Data Assimilation in Latent Space}
\noindent \textbf{Forward diffusion}. In the forward diffusion process, a sample $x(t) \sim p(x)$ is progressively perturbed through a continuous-time diffusion process expressed as linear stochastic differential equation (SDE) as in Equation \ref{eq:forward} \cite{song2020score},
\begin{equation}
    \label{eq:forward}
    dx(t) = \underbrace{f(t)x(t) dt}_{\text{drift term}} + \underbrace{g(t)dw(t)}_{\text{diffusion term}},
\end{equation}

\noindent where $f(t), g(t) \in \mathbb{R}$ are the drift and diffusion coefficients, $w(t) \in \mathbb{R}^\mathcal{L}$ the standard Brownian motion (Wiener process), and $x(t) \in \mathbb{R}^\mathcal{L}$ the perturbed sample at time $t \in [0,1]$. Since the source of randomness from the Wiener process is Gaussian, a linear SDE w.r.t. $x(t)$ ensures that the perturbation kernel is also Gaussian with the form $p(x(t) \mid x) = \mathcal{N}(\mu(t)x, \Sigma(t))$ \cite{sarkka2019applied}. As a result, $\mu(t) \text{ and } \Sigma(t)$ can be derived analytically from $f(t), g(t)$ \cite{sarkka2019applied}. 

Furthermore, in a standard diffusion process, we want our sample distribution at $t = 0$ to be as close to the underlying data distribution $p(x)$ as possible, such that $p(x(0)) \approx \mathcal{N}(x, 0)$. This means that $\mu(0) = 1 \text{ and } \Sigma(0) \ll 1$. On the other hand, we want our sample distribution at $t = 1$ to be as close to Gaussian noise as possible, such that $p(x(1)) \approx \mathcal{N}(0, \Sigma(1))$, implying $\mu(1) = 0$. There are multiple SDEs that satisfy these boundary and evolution constraints, including variance-exploding (VE) and variance-preserving (VP) SDEs \cite{song2020score}.\\

\noindent \textbf{Reverse denoising}. The reverse denoising process is represented by a reverse SDE as defined in Equation \ref{eq:reverse} \cite{song2020score}. Notice that the drift and diffusion terms are similar to those found in the forward SDE, and that the only difference is the \emph{score}. If we have access to the true score, we can perfectly recover the original data samples from noise as $t$ from 1 to 0. Intuitively, the score guides the reverse process in the direction where the probability of observing $x(t)$ is highest.
\begin{equation}
    \label{eq:reverse}
    \begin{aligned}
    dx(t) &= [\underbrace{f(t)x(t)}_{\text{drift term}} - g(t)^2 \underbrace{\nabla_{x(t)}\log p(x(t))}_{\text{score}}] dt \\
            & + \underbrace{g(t)dw(t)}_{\text{diffusion term}}.
    \end{aligned}
\end{equation}

\noindent \textbf{Training score function}. In practice, the score is approximated with a neural network called the score network, $s_{\phi}(x(t), t)$. The objective function would then be to minimize a continuous weighted combination of Fisher divergences between $s_{\phi}(x(t), t)$ and $\nabla_{x(t)}\log p(x(t))$ through score matching \cite{song2020score, vincent2011connection}. However, the perturbed data distribution $p(x(t))$ is complex, data-dependent, and hence unscalable. As such, \cite{vincent2011connection} reformulated the objective function by replacing the score with the perturbation kernel where we have access to the analytical form as discussed earlier (Equation \ref{eq:score_matching}; expectation is taken over $p(x), p(t), \text{ and } p(x(t) \mid x)$).
\begin{equation}
    \label{eq:score_matching}
    \arg \min_{\phi} \mathbb{E}\left[ \Sigma(t)  \left\| s_{\phi}(x(t),t) - \nabla_{x(t)}\log p(x(t) \mid x) \right\|^2_2 \right].
\end{equation}

Several studies have noted the instability of this objective function as $t \rightarrow 0$, so \cite{zhang2022fast} suggests a reparameterization trick which replaces $s_{\phi}(x(t),t) = \epsilon_{\phi}(x(t),t) / \sigma(t)$, where $\Sigma = \sigma^2$ (Equation \ref{eq:score_matching_reparam}; expectation is taken over $p(x), p(t), p(\epsilon) \sim \mathcal{N}(0, I)$).
\begin{equation}
    \label{eq:score_matching_reparam}
    \arg \min_{\phi} \mathbb{E}\left[ \left\| \epsilon_{\phi}(\mu(t)x + \sigma(t)\epsilon, t) - \epsilon) \right\|^2_2 \right].
\end{equation}

\noindent Following convention used by previous works on score-based diffusion model, we will denote $\epsilon_{\phi}(x(t),t)$ with $s_{\phi}(x(t),t)$ for cleaner notation.\\

\noindent \textbf{Conditioning the generative process}. The case we have discussed so far is the unconditional generative process as we try to sample $x \sim p(x(0))$ from a prior noise distribution. In order to condition the generative process with observations $y$, we seek to sample from $x \sim p(x(0) \mid y)$. This can be done by modifying the score as in Equation \ref{eq:reverse} with $\nabla_{x(t)}\log p(x(t) \mid y)$ and plugging it back to the reverse SDE process. 

However, one would need re-training whenever the observation process, $p(y \mid x)$, changes. Nonetheless, several works have attempted to approximate the conditional score with just a single pre-trained score network, eliminating the need for expensive re-training \cite{song2020score, chung2022diffusion}.

First, using Bayes rule, we expand the conditional score as in Equation \ref{eq:score_bayes},
\begin{equation}
    \label{eq:score_bayes}
    \begin{aligned}
    \nabla_{x(t)}\log p(x(t) \mid y) &= \underbrace{\nabla_{x(t)}\log p(x(t))}_{\text{unconditional score}}\\
    & + \underbrace{\nabla_{x(t)}\log p(y \mid x(t))}_{\text{likelihood score}}.
    \end{aligned}
\end{equation}
Since the first term on the right-hand side is already approximated by the score function, the remaining task is to identify the second likelihood score. Assuming differentiable measurement function $\mathcal{A}$, and a Gaussian observation process in the latent space, $p(y \mid x(t)) = \mathcal{N}(\mathcal{A}(x(t)), \sigma^2_y)$, the approximation goes as in Equation \ref{eq:likelihood_approx} \cite{chung2022diffusion},
\begin{equation}
    \label{eq:likelihood_approx}
    \begin{aligned}
    p(y|x(t)) = \int p(y \mid x) p(x \mid x(t)) dx\\
    \simeq \mathcal{N}(y \mid \mathcal{A}(\hat{x}(x(t)), \sigma^2_y),
    \end{aligned}
\end{equation}

\noindent where $\hat{x}(x(t))$ can be approximated by the Tweedie's formula \cite{efron2011tweedie} as in Equation \ref{eq:tweedies},
\begin{equation}
    \label{eq:tweedies}
    \hat{x}(x(t)) \simeq \frac{x(t) + \sigma^2(t)s_{\phi}(x(t), t)}{\mu(t)}.
\end{equation}

\noindent The implication is that we can now approximate $\nabla_{x(t)}\log p(x(t) \mid y)$ with just a singly trained score network, allowing for a zero-shot assimilation where we do not require constant finetuning when distributions shift.

\subsection{Scalability and Numerical Stability}
We also implemented several performance and stability improvements as proposed in \cite{rozet2023score}. For brevity, we leave out the theoretical details and refer interested reader to \cite{rozet2023score}. 

These improvements include using a subset of embedded trajectories $\{x_i, x_{i+1}, \cdots, x_{i+K-1}\} \in x_{1:T}$ to approximate the entire sequence during training, termed as a Markov blanket (throughout this work, we set $K = 5$). Practically, this strategy helps to reduce the computation cost in cases where the sequence length grows prohibitively large, as is commonly found in several Earth system processes. Also, we reparameterize Equation \ref{eq:likelihood_approx} to that in Equation \ref{eq:likelihood_reparam} to improve the numerical instability of $\hat{x}(x(t))$ due to the division by $x(t)$ in Equation \ref{eq:tweedies},
\begin{equation}
    \label{eq:likelihood_reparam}
    \begin{aligned}
    p(y|x(t)) \simeq \mathcal{N}(y \mid \mathcal{A}(\hat{x}(x(t)), \sigma^2_y + \underbrace{\gamma\frac{\sigma^2(t)}{\mu^2(t)}I}_{\text{stability term}}),
    \end{aligned}
\end{equation}

\noindent where $\gamma$, $I$ are scalar constant and identity matrix respectively. This re-parameterization trick reduces numerical instability especially at the beginning of the reverse denoising process where $\mathbb{E}[x(t)] \rightarrow 0$ as $t \rightarrow 1$. Intuitively, the stability term ensures that the likelihood approximation is adjusted according to the noise-signal ratio (i.e., $\sigma^2 / \mu^2$): a noisier state, $x(t)$, should result in a more diffused likelihood approximation, and vice versa.

Last, we implement a predictor-corrector procedure to improve the quality of our conditional generative process. The reverse SDE \emph{prediction} process is solved using the exponential integrator (EI) discretization scheme \cite{zhang2022fast} as in Equation \ref{eq:prediction}, and the \emph{correction} uses a few steps of Langevin Monte Carlo (LMC) as in Equation \ref{eq:correction},
\begin{equation}
    \label{eq:prediction}
    \begin{aligned}
    x(t - \Delta t) &\leftarrow \frac{\mu(t - \Delta t)}{\mu(t)}x(t) \\
    & + \left(\frac{\mu(t - \Delta t)}{\mu(t)} - \frac{\sigma(t - \Delta t)}{\sigma(t)}\right)\Sigma(t)s_{\phi}(x(t), t),
    \end{aligned}
\end{equation}

\begin{equation}
    \label{eq:correction}
    x(t) \leftarrow x(t) + \tau s_{\phi}(x(t), t) + \sqrt{2\tau}\epsilon.
\end{equation}

\noindent The correction process is important because it corrects sample generation using information from the gradient of $\log p(x(t))$ (approximated by $s_{\phi}(x(t), t)$), which guides the samples towards regions of higher probability, and adds randomness modulated by the Langevin amplitude step $\tau$.

\section{Experimental Setup}
\label{sec:setup}

\subsection{Datasets}
A standard DA framework generally attempts to calibrate \emph{model states} with available \emph{observations}. The former typically comes from a more homogeneous model output with intrinsic uncertainty, while the latter is more heterogeneous in terms of its modality, resolution, and quality. For this work, we consider two major observation modalities: \emph{in-situ} weather stations and \emph{ex-situ} satellite imagery. 

\begin{itemize}
    \item \emph{Model states}. We use ERA5 reanalysis as our approximation to the true states \cite{hersbach2020era5}. ERA5 reanalysis was produced by the European Centre for Medium-Range Weather Forecasts (ECMWF) that recalibrates their historical forecasts or hindcasts with available observations using state-of-the-art DA method. We select temperature across 10 pressure levels, such that $t@\{10,50,100,200,300,500,700,850,925,1000\}$-hpa, following specifications suggested by \cite{nathaniel2024chaosbench}. 

    \item \emph{In-situ observations}. We use global rain gauges data provided by the Climate Prediction Center (CPC) which has daily resolution and dates back to 1979 \cite{chen2008assessing}. Precipitation is an important phenomenon that captures complex interaction on land, ocean, and atmosphere \cite{manabe1964thermal, manabe1967thermal} and which is challenging to model by Earth system models.

    \item \emph{Ex-situ observations}. We use outgoing longwave radiation (OLR) data collected on-board the NOAA-14, NOAA-16, and NOAA-18 satellites \cite{liebmann1996description}. OLR is an important parameter that is closely related to the planetary energy budget where temperature profile is coupled \cite{manabe1964thermal}.
\end{itemize}

\noindent We resample all datasets to a 1.5-degree grid at daily resolution. We divide the years 2000-2015 for training, 2016-2021 for validation, and 2022 for testing. We refer to (i) \emph{unimodal} and (ii) \emph{multimodal} as models that incorporate only background state or also take into account \emph{in-situ} and \emph{ex-situ} observations for conditioning. The measurement function $\mathcal{A}$ is assumed to be differentiable. If this is not the case, a deep-learning emulation of the physical process can be used so that it would be differentiable. Non-latent approaches (i.e., pixel-based) are by construction \emph{unimodal} due to the absence of $\mathcal{H}$ to reconcile the heterogeneous state-observation spaces discussed earlier. 

\subsection{Details on Autoencoder}
We use a convolution-based encoder-decoder structure with 5 layers, ReLU activation \cite{nair2010rectified}, kernel size of 5, stride of 1, and a batch normalization \cite{ioffe2015batch}. We train over 64 epochs, optimized with \textsc{AdamW} \cite{loshchilov2017decoupled}, with a batch size of 64, and follows a linearly decaying scheduler with an initial learning rate of $2\times10^{-4}$ and a weight decay of $1\times10^{-3}$. 


\subsection{Details on Score Network}
We use U-Net \cite{ronneberger2015u} with residual blocks \cite{he2016deep}, SiLU activation \cite{elfwing2018sigmoid}, and layer normalization \cite{ba2016layer} for our score network. In particular, the network consists of 3 layers consisting of [3, 3, 3] residual blocks. We specify the hidden channels to be [64, 128, 256] and [32, 64, 128] for either the pixel-based or latent-based score network, with the latter approach enjoying a 8x reduction in model size. Furthermore, we use a circular padding to ensure continuous representation of a spherical globe \cite{islam2021position}. We train over 64 epochs, optimized with \textsc{AdamW} \cite{loshchilov2017decoupled}, with a batch size of 64, and follows a linearly decaying scheduler with an initial learning rate of $2\times10^{-4}$ and a weight decay of $1\times10^{-3}$. Finally, we introduce a two-layer linear network to represent the temporal component of the score network with a hidden and embedding size of 256 and 64 respectively. 

During inference, we apply 512 denoising steps with 1-step LMC correction, $\gamma=10^{-2}$, and $\tau=0.5$. 
\section{Results and Discussion}
\label{sec:results}
In this section, we present an extensive ablation study evaluating the consistency of our proposed latent-based framework, \texttt{SLAMS}, when compared to its pixel-based counterpart. We divide our discussion into the (i) ideal case and (ii) realistic scenario of data assimilation, depending on the quality of conditioning inputs. In addition, we perform multimodal feature importance study to understand the contribution of different observation modality. 

\begin{figure}
    \centering
    \includegraphics[width=0.45\textwidth]{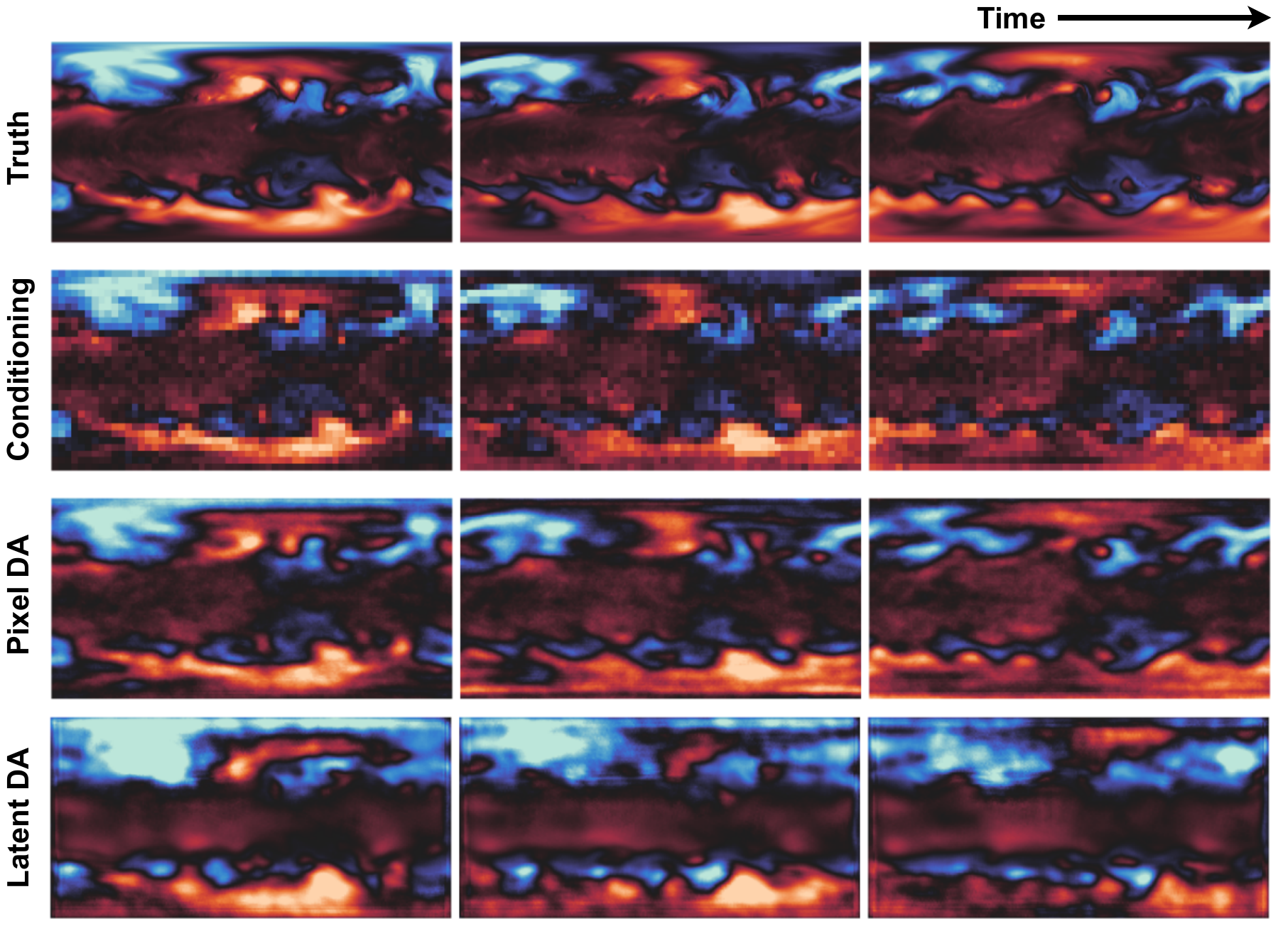}
    \caption{Ideal case where we have \textbf{high resolution, low noise, dense inputs} (4x coarsening, $\sigma^2=0.1$) to calibrate $t@200hpa$. We find that pixel-based DA generates qualitatively better assimilated states.}
    \label{fig:high_quality_obs}
\end{figure}

\begin{figure}
    \centering
    \includegraphics[width=0.45\textwidth]{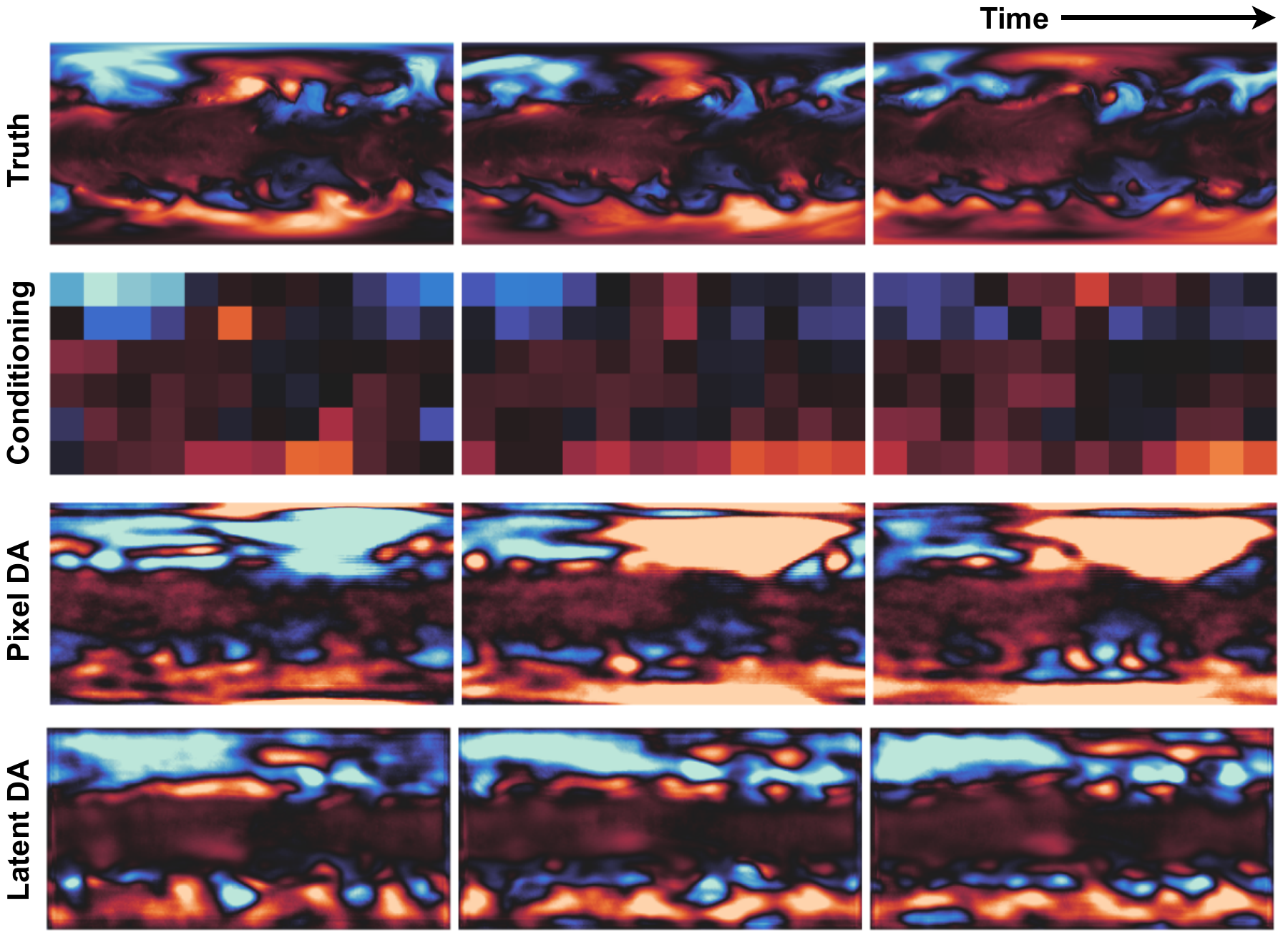}
    \caption{Realistic case where we have \textbf{low-resolution inputs} (20x coarsening) to calibrate $t@200hpa$. Our latent-based DA approach, \texttt{SLAMS}, is physically more consistent.}
    \label{fig:low_quality_obs_coarse}
\end{figure}

\subsection{Ideal Case: Assimilating High Quality Data}
We first begin our discussion in an idealized case where we are given high-resolution, low noise, and dense inputs. We mimic this scenario by applying a coarsening factor of 4, additive Gaussian noise with $\sigma^2 = 0.1$, and no masking.

As illustrated in Figure \ref{fig:high_quality_obs}, we find that pixel-based DA approach generates qualitatively better assimilated states when compared to our latent-based approach. This is akin to introducing additional small noises into the reverse SDE process, and a well-trained score network is robust to denoise such slight perturbation. 

\subsection{Realistic Case: Assimilating Low Quality Data}
Unfortunately, many real-world scenarios tend to provide us with (i) low-resolution, (ii) noisy, and (iii) sparse data. We attempt to replicate these scenarios by applying extreme coarsening factor of 20, additive Gaussian noise of $\sigma^2=4.0$, or sampling gap of 16, respectively.

\begin{figure}
    \centering
    \includegraphics[width=0.45\textwidth]{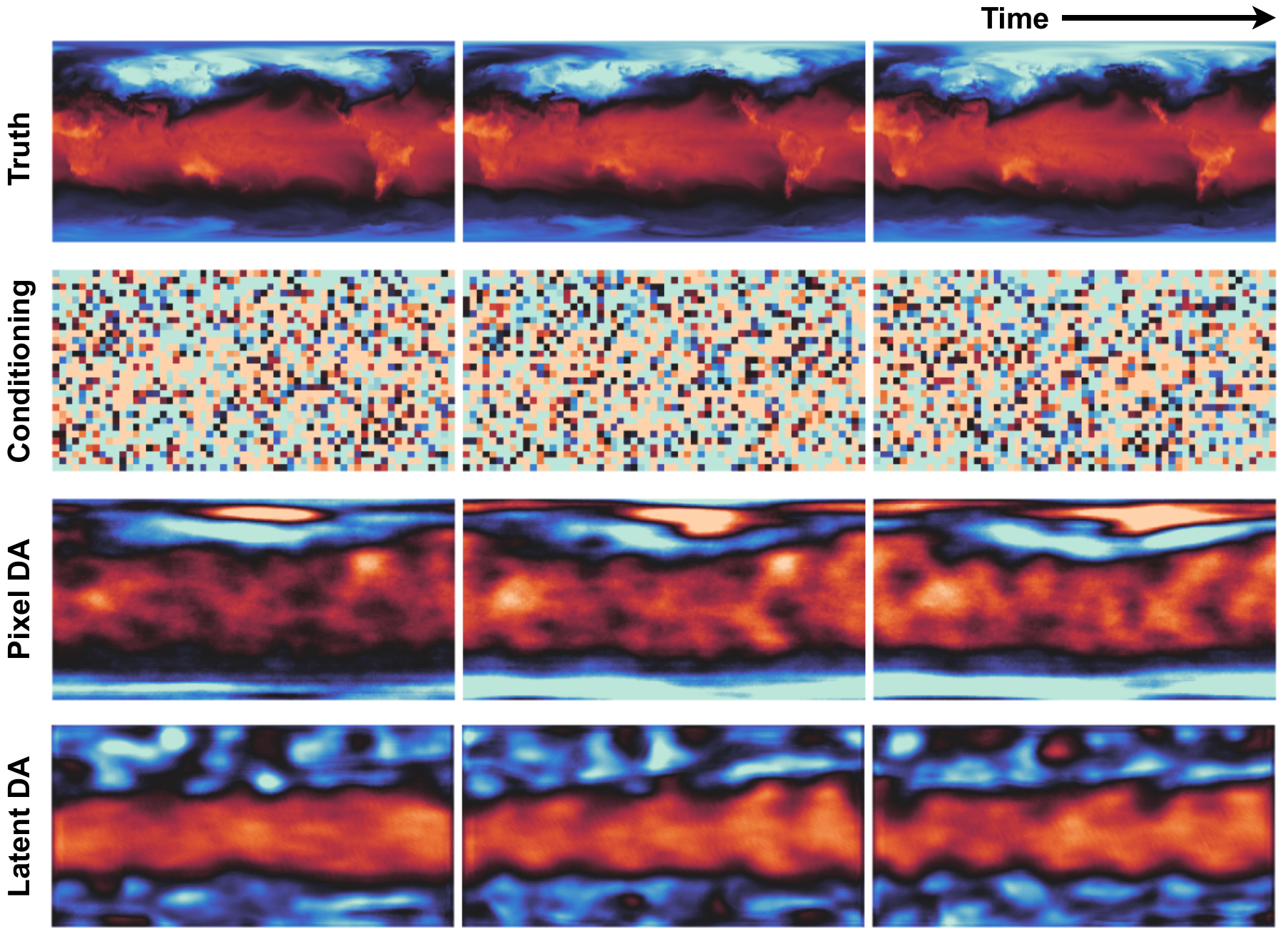}
    \caption{Realistic case where we have \textbf{noisy inputs} ($\sigma^2 = 4$) to calibrate $t@1000hpa$. Our latent-based DA approach, \texttt{SLAMS}, is physically more consistent.}
    \label{fig:low_quality_obs_noisy}
\end{figure}

\begin{figure}
    \centering
    \includegraphics[width=0.45\textwidth]{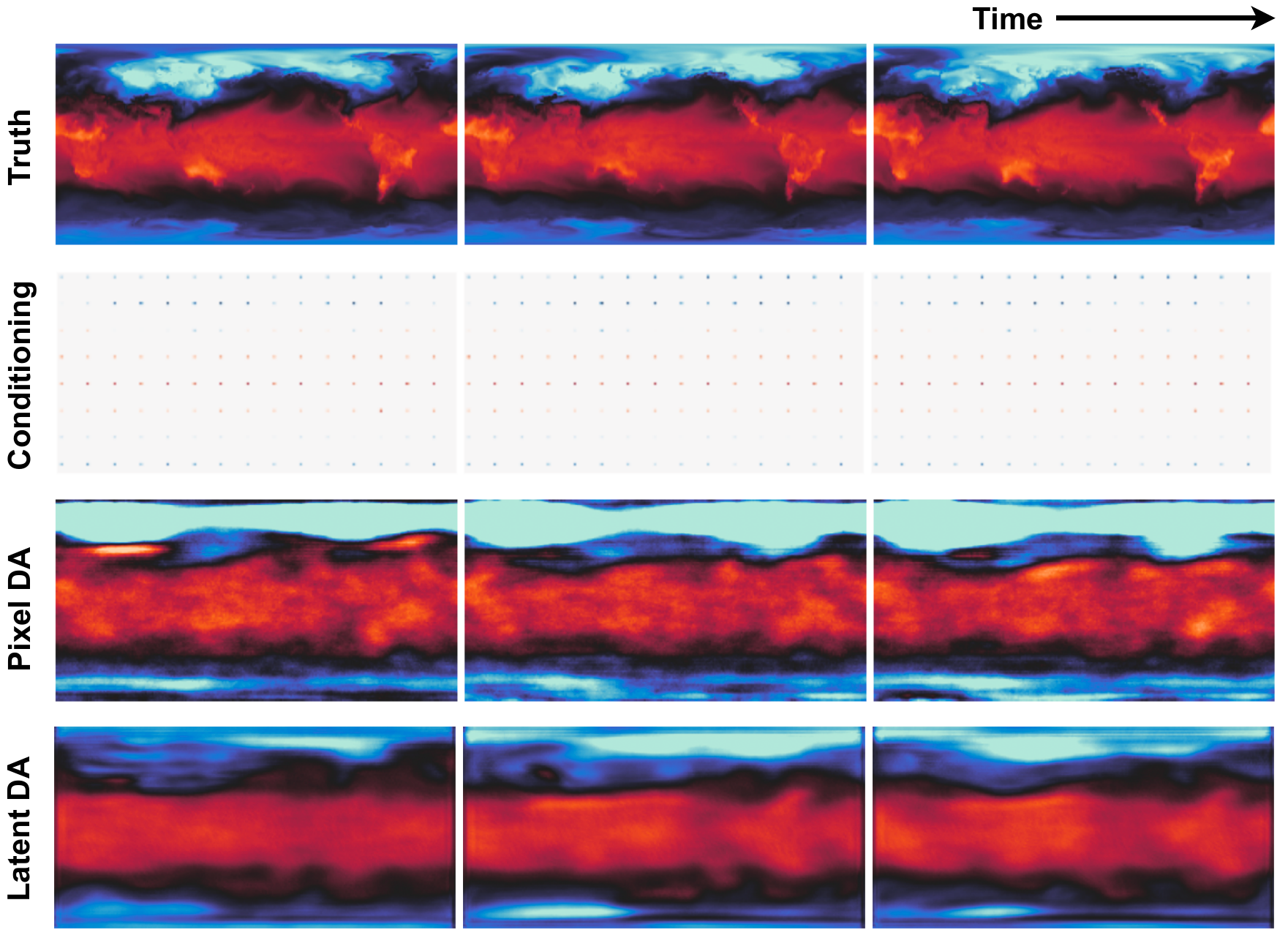}
    \caption{Realistic case where we have \textbf{sparse inputs} (16x sampling gap) to calibrate $t@1000hpa$. Our latent-based DA approach, \texttt{SLAMS}, is physically more consistent.}
    \label{fig:low_quality_obs_sparse}
\end{figure}

\noindent \textbf{Low-resolution data}. As illustrated in Figure \ref{fig:low_quality_obs_coarse}, we find that our latent-based DA approach tends to be temporally and physically consistent especially in high latitudes where we do not observe unrealistic abrupt cooling $\rightarrow$ warming changes, which are otherwise observed in the pixel-based DA approach.

\noindent \textbf{Noisy data}. As illustrated in Figure \ref{fig:low_quality_obs_noisy}, we find that our latent-based DA approach produces more realistic continuous pattern in the tropics. Compared to the pixel-based approach, \texttt{SLAMS} also generates samples with more accurate physics, with no unrealistic large-scale near-surface warming in the polar region.

\noindent \textbf{Sparse data}. As illustrated in Figure \ref{fig:low_quality_obs_sparse}, we find that our latent-based DA approach does not inaccurately overestimate near-surface temperature in the high-latitude, and generate realistic, smoother calibration in the tropics.

These results suggest the stabilizing capability of our latent-based approach, \texttt{SLAMS}, that can be partly explained by: (i) reduced dimensionality of the latent code that makes it less sensitive to high-dimensional, pixel-level perturbation \cite{zhao2022defeat}, (ii) stabilizing characteristic of a well-trained decoder exhibiting posterior non-collapse \cite{rombach2022high}, (iii) additional constraints enforced by increasing data modality \cite{fei2022towards}.

\begin{figure*}[h!]
    \centering
    \begin{subfigure}{0.95\textwidth}
        \includegraphics[width=\textwidth]{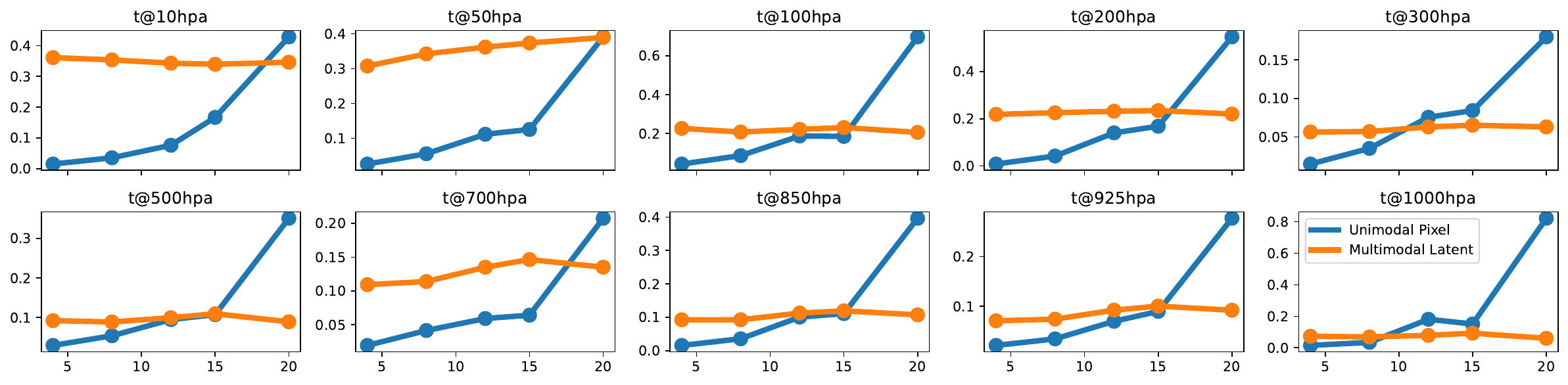}
        \caption{Wasserstein distance against coarsening factors. Our multimodal latent-based DA approach (\texttt{SLAMS}) is consistent even when coarsening factors $\geq 15$}
    \end{subfigure}
    \hfill
    
    \begin{subfigure}{0.95\textwidth}
        \includegraphics[width=\textwidth]{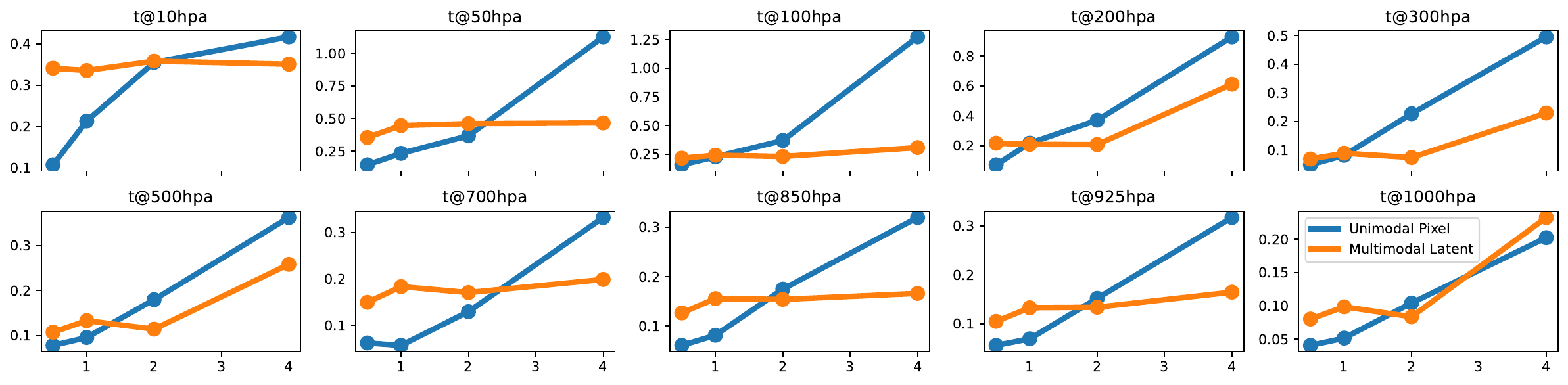}
        \caption{Wasserstein distance against noise level. Our multimodal latent-based DA approach (\texttt{SLAMS}) is consistent even when $\sigma^2 \geq 2$}
    \end{subfigure}
    \hfill

    \begin{subfigure}{0.95\textwidth}
        \includegraphics[width=\textwidth]{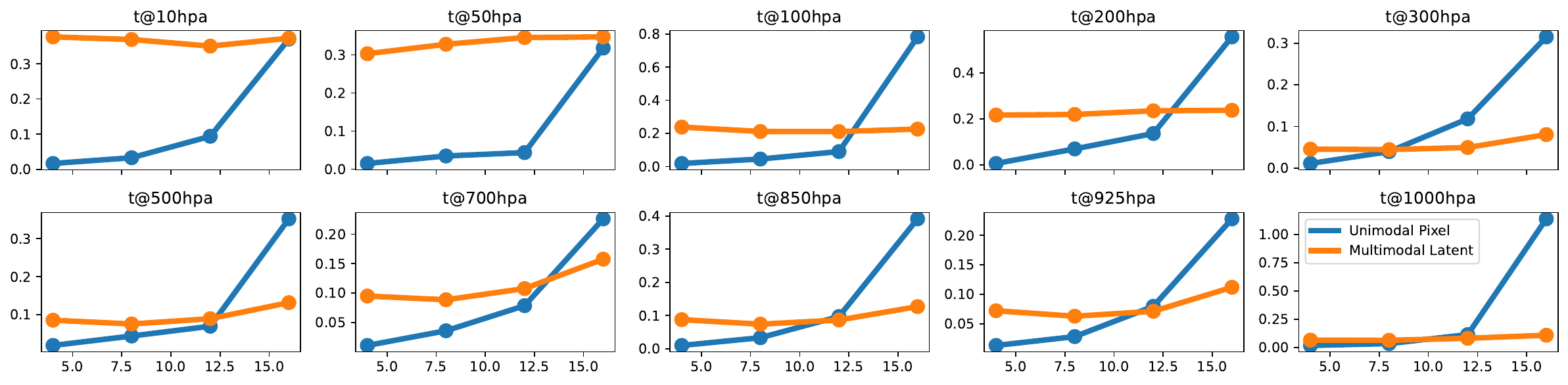}
        \caption{Wasserstein distance against sampling gap. Our multimodal latent-based DA approach (\texttt{SLAMS}) is consistent even when sampling gap $\geq 12$}
    \end{subfigure}
    
    \caption{Ablation result evaluating the effects of varying (a) resolution, (b) noise level, and (c) sparsity on Wassterstein distance ($\downarrow$ is better) of different diffusion-based DA methods. Across most target states, a diffusion-based DA approach operating in latent space (\texttt{SLAMS}) is more stable than that in the pixel space beyond certain threshold.}
    \label{fig:ablation}
\end{figure*}

\begin{figure*}[h!]
    \centering
    \begin{subfigure}{0.95\textwidth}
        \includegraphics[width=\textwidth]{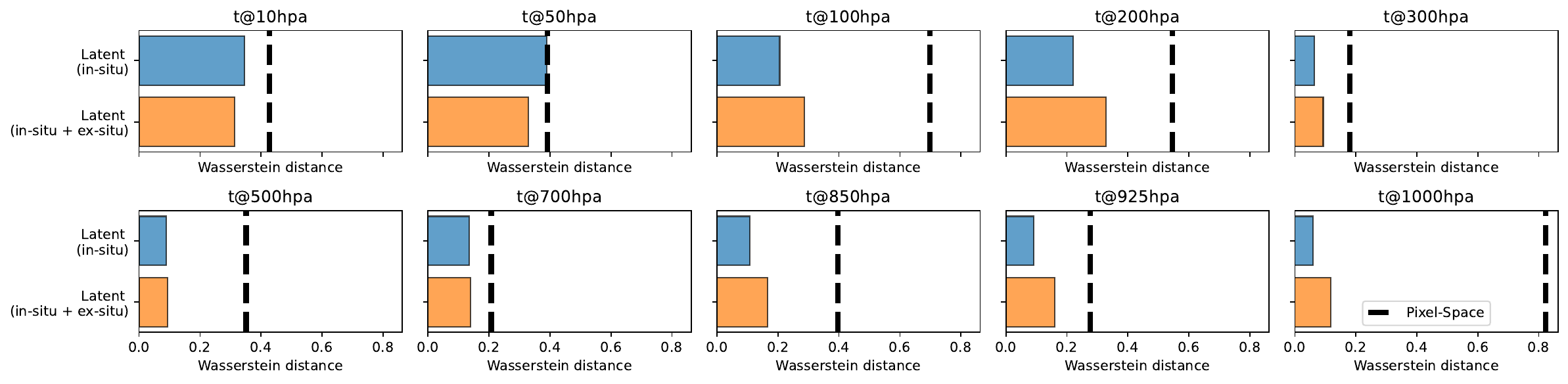}
        \caption{Effect of assimilating low-resolution multimodal inputs (20x coarsening factor)}
    \end{subfigure}
    \hfill
    
    \begin{subfigure}{0.95\textwidth}
        \includegraphics[width=\textwidth]{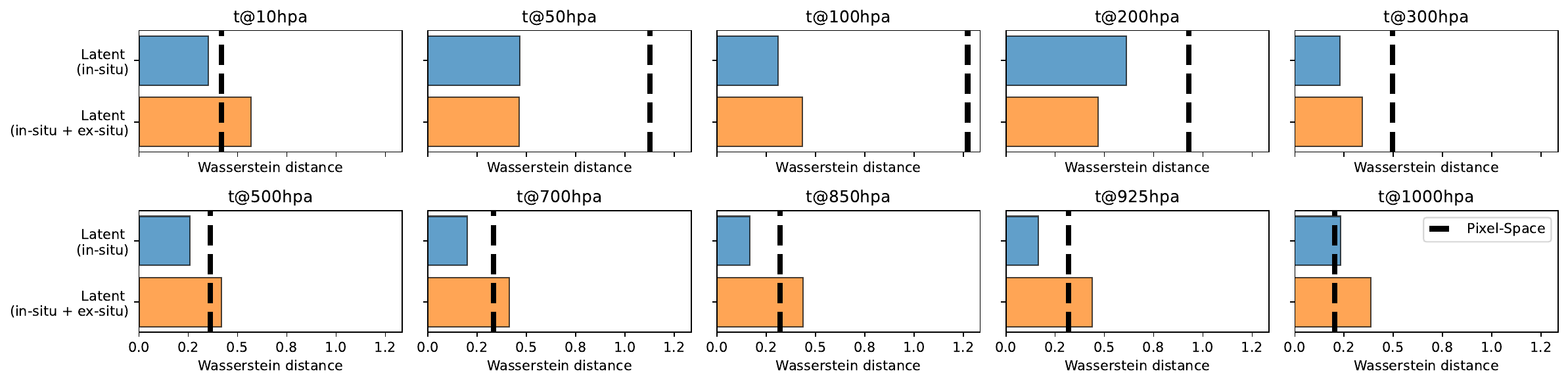}
        \caption{Effect of assimilating noisy multimodal inputs ($\sigma^2 = 4$)}
    \end{subfigure}
    \hfill

    \begin{subfigure}{0.95\textwidth}
        \includegraphics[width=\textwidth]{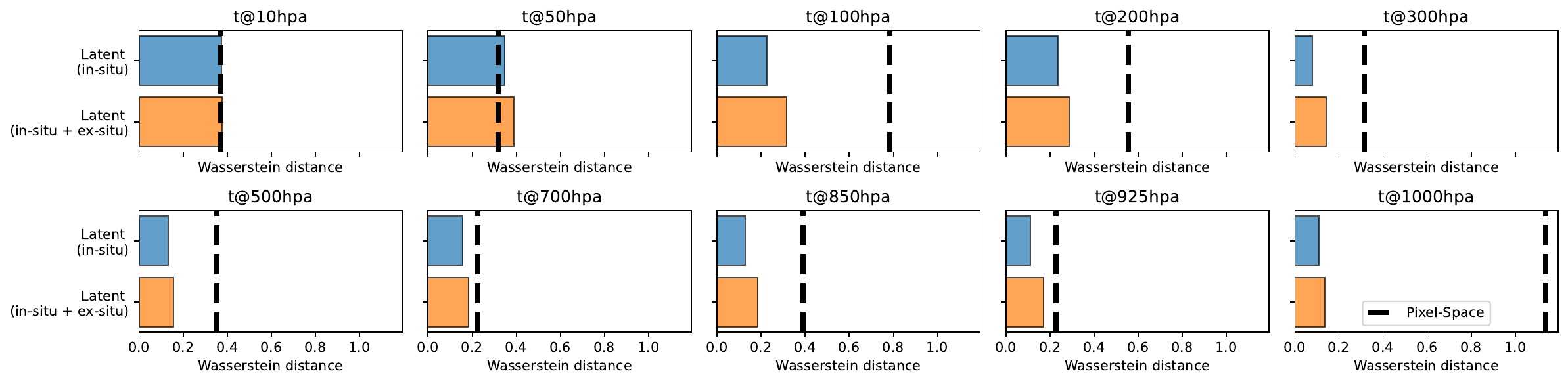}
        \caption{Effect of assimilating sparse multimodal inputs (16x sampling gap)}
    \end{subfigure}
    
    \caption{Ablation result on the effect of multimodality in the calibration of target states given (a) low-resolution, (b) noisy, and (c) sparse inputs. The x-axis shows the Wasserstein distance ($\downarrow$ is better). The results demonstrate that adding ex-situ imagery (orange bar), mostly enhances top-of-atmosphere state calibration (e.g., $t@10hpa, t@50hpa$) compared to only conditioning on in-situ observations (blue bar). Both latent methods (\texttt{SLAMS}) outperform pixel-based approach (vertical dashed line) in most target states.}
    \label{fig:feature}
\end{figure*}

\subsection{Toward Consistent Multimodal Assimilation}
In order to fully showcase the robustness of our latent-based approach, \texttt{SLAMS}, we conducted an in-depth ablation study that varied the coarsening factor, noise level, and sampling gap given (i) \emph{unimodal} (only background), or (ii) \emph{multimodal} (background + observations) conditioning inputs. Specifically, we employed the Wasserstein distance \cite{monge1781memoire} to quantify the distribution divergence between the true and assimilated states. Full results are illustrated in Figure \ref{fig:ablation}. 

Overall, our ablation study indicates that \texttt{SLAMS} maintains its robustness (i.e., exhibits lower Wasserstein distances) across the majority of target states, particularly when assimilating inputs that are of low resolution (coarsening factors $\geq 15$), highly noisy ($\sigma^2 \geq 2$), and sparse (sampling gap $\geq 12$).

First, an operationally useful grid resolution for Earth system model typically falls below 0.1 degrees \cite{o2016scenario}. However, many datasets such as those used in our study are generally an order of magnitude coarser, approximately 1.5 to 2.5 degrees. This discrepancy can lead to instability and sensitivity to perturbations in pixel-based DA approaches.

Likewise, the challenge of noisy observations, compounded by model uncertainty, is a widespread issue. Traditional DA systems must account for these through expensive error propagation calculations, which can become intractable in multimodal, high-dimensional scenarios \cite{janjic2018representation}. Thus, a robust DA system, underpinned by a deep probabilistic framework like ours, has to be robust to noise.

Lastly, data sparsity is a common challenge in DA for Earth system modeling. Sparsity can come in many forms, including spatial (swath width, coverage), temporal (orbit period), and/or representativeness (geostationary orbit) \cite{nathaniel2023metaflux}. Thus, a DA framework that can efficiently extract signals is crucial for robust real-time modeling.

\subsection{Multimodal Feature Ablation} 
So far, we have demonstrated the feasibility of performing DA in a multimodal setting through a unified representational space. The natural next step is to assess the impact of multimodal observations in improving the quality of analysis states. Figure \ref{fig:feature} reaffirms our earlier findings that our latent-based DA approach, \texttt{SLAMS}, outperforms pixel-based DA across most target states. Additionally, the inclusion of ex-situ imagery is particularly valuable for constraining top-of-atmosphere (ToA) variables (e.g., $t@10hpa, t@50hpa$). This finding is consistent with the role of OLR as a proxy for the planetary energy budget \cite{huang2007investigation}.

One strategy to maximize the extraction of signals from multimodal observations involves accounting for different latent representations of various data modalities. Future work could look into distinct, more powerful autoencoders (e.g., VQ-VAE \cite{van2017neural}) for each modality and introducing an aggregator to boost the expressiveness of each feature \cite{cheng2023generalised}. This approach also provides the benefit of making the framework adaptable to different underlying data structures that are not easily gridded as image frames, including LiDAR point clouds, textual, and tabular data.
\section{Conclusion}
\label{sec:conclusion}
We introduce \texttt{SLAMS}, a score-based latent data assimilation framework that enables multimodality. Our framework reframes the traditionally compute-intensive DA process using  deep probabilistic principles through diffusion as a powerful class of generative model. \texttt{SLAMS} facilitates the assimilation of multimodal data through its latent-centric architecture, capitalizing on recent advancements in dimensionality reduction and automatic feature extraction. The adaptable design of \texttt{SLAMS} also accommodates future integration with data modalities traditionally challenging to represent as image frames, such as point clouds. Moreover, the probabilistic underpinning of our framework allows for ensemble generation of analysis, enabling uncertainty quantification. We validate \texttt{SLAMS}'s efficacy in realistic scenario by assimilating multimodal \emph{in-situ} weather station data and \emph{ex-situ} satellite imagery to calibrate the vertical temperature profile, globally. Extensive ablation confirms that \texttt{SLAMS} is robust to low-resolution, noisy, and sparse inputs. To our knowledge, this work is the first to propose a data-driven \emph{probabilistic, multimodal} data assimilation framework for real-world Earth system modeling.
\section*{Acknowledgments}
The authors would like to acknowledge funding from the National Science Foundation Science and Technology Center, Learning the Earth with Artificial intelligence and Physics, LEAP (Grant number 2019625), Department of Energy grant.

\

{
    \small
    \bibliographystyle{ieeenat_fullname}
    \bibliography{main}
}


\end{document}